%% file: main_v3.tex
\documentclass{article}
\usepackage[T1]{fontenc}
\usepackage{spconf,amsmath,amssymb,graphicx,booktabs,array,tabularx,balance}
\usepackage[hidelinks]{hyperref}
\usepackage{xcolor}
\usepackage{orcidlink}
\hypersetup{
pdftitle={Learning Multimodal Sensor Fusion with JEPA Representations for Breathing Activity and Pose Recognition},
pdfauthor={Yunxiang Luo et al.},
pdfsubject={Controlled multimodal classification with frozen JEPA features and missing sensors}
}

\graphicspath{{figures/}}
\newcommand{\modalities}{\mathcal{M}}
\newcommand{\visible}{S}
\newcommand{\hrep}{\mathbf{h}}
\newcommand{\zrep}{\mathbf{z}}
\newcommand{\stopgrad}{\operatorname{sg}}
\newcommand{\lossCJ}{\mathcal{L}_{\mathrm{cj}}}

\title{Learning Multimodal Sensor Fusion with JEPA Representations \\ for Breathing Activity and Pose Recognition}

\name{\begin{tabular}{c}Yunxiang Luo$^{\dagger}$$^{\ast}$$^{\orcidlink{0009-0003-2750-8074}}$,
Constantino {\'A}lvarez Casado$^{\dagger}$$^{\ddagger}$$^{\ast}$$^{\orcidlink{0000-0002-3052-4759}}$,
Xiaoting Wu$^{\dagger}$$^{\orcidlink{0000-0003-2019-2874}}$, Nhi Nguyen$^{\dagger}$$^{\orcidlink{0009-0002-2090-0746}}$,
\\
Le Nguyen$^{\dagger}$$^{\orcidlink{0000-0001-7765-1483}}$, Manuel Lage Ca{\~n}ellas$^{\dagger}$$^{\ddagger}$$^{\orcidlink{0000-0002-4917-340X}}$,
Miguel Bordallo L{\'o}pez$^{\dagger}$$^{\orcidlink{0000-0002-5707-9085}}$\end{tabular}}

\address{
$^{\dagger}$Center for Machine Vision and Signal Analysis, University of Oulu, Finland \\
$^{\ddagger}$Candour Ltd, Oulu, Finland \\
\thanks{$^{\ast}$These authors contributed equally to this work.}
}

\begin{document}
\ninept
\maketitle

\begin{abstract}
Predictive video representations do not by themselves specify how a multimodal system should handle missing sensors. We transfer frozen V-JEPA2 features to a set of heterogeneous sensors comprised by RGB video, depth video, and radar time series. A shared fusion model uses modality queries, fixed sensor slots, and explicit availability to classify breathing activity and pose from all seven non-empty sensor subsets. The contribution is a controlled evaluation of subset supervision, pooling capacity, and representation transfer on 39 OMuSense-23 subjects with five subject-group folds and three seeds. With matched downstream parameter counts, the final model reaches recording-level activity macro-F1 of $75.23\%\pm0.60$ averaged across seven subsets. Supervising all subsets in each batch improves the matched query-pooling control by 1.923 percentage points (conditional 95\% subject-bootstrap interval [0.647, 3.209]), while query pooling shows no clear advantage over matched mean pooling under balanced supervision. Radar probes support transfer beyond pooled wavelet pixels, 64-token inference approximately preserves mean activity performance, and the tested Consensus-JEPA loss adds no measured benefit against its exact control.
\end{abstract}


\section{Introduction}
\begin{figure}[!t]
\centering
\includegraphics[width=\columnwidth]{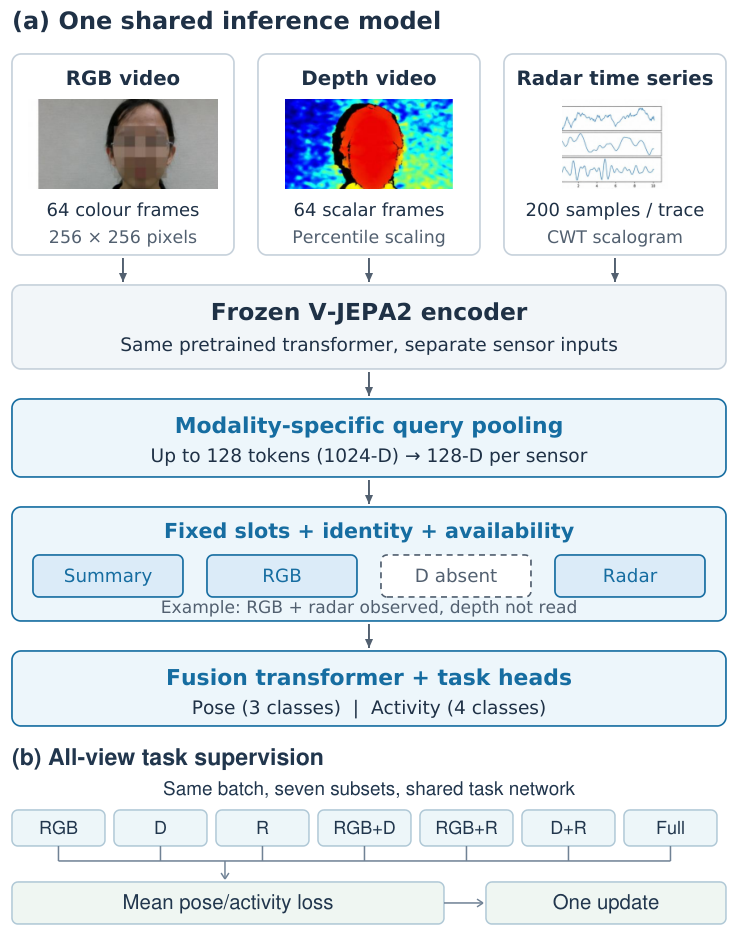}
\vspace{-6mm}
\caption{(a) Shared inference architecture of the final query model and the earlier Subset-query (B3). Grey marks frozen feature extraction and blue trained pooling and fusion. The example declares depth absence, and insets show representative observations. (b) Final all-view task supervision evaluates one batch under all seven non-empty subsets, averages the pose/activity losses, and makes one update, without a latent prediction loss.}
\label{fig:architecture}
\vspace{-1mm}
\end{figure}

Multimodal contactless human sensing combines cameras and radar to extract physiological and behavioural information for health monitoring and human state assessment~\cite{omusense2026}. RGB, depth, and radar observe the appearance, geometry, and motion of the same activity and can provide complementary information about breathing and body pose, but the available sensor configuration may change between data collection and deployment.

Power, synchronization, and communication requirements can restrict which sensors are deployed~\cite{de2020input}. Here, a missing modality denotes an unavailable sensor stream, rather than degradation within a stream that remains available. Modality dropout and adaptation of pretrained networks address this condition~\cite{de2020input,reza2024robust}, yet multimodal transformers remain sensitive to missing inputs~\cite{ma2022multimodal}. Performance with complete inputs therefore does not establish performance with missing sensors.

We use V-JEPA2, a video model trained with a joint-embedding predictive architecture (JEPA) to predict target representations from available context~\cite{assran2025vjepa2}. The central question is how downstream pooling and supervision affect the use of these frozen representations for breathing activity and pose classification when sensor availability changes. The contributions are as follows:
\begin{enumerate}
\setlength{\itemsep}{1pt}
\setlength{\parskip}{0pt}
\item \textbf{Availability-aware multimodal fusion.} One classifier with query pooling, fixed sensor slots, and explicit availability covers all seven non-empty RGB, depth, and radar subsets, reaching $75.23\pm0.60$\% activity macro-F1 averaged across them.
\item \textbf{Controlled attribution of training gains.} With matched downstream parameters, supervising all subsets in each batch improves the query model by 1.923 percentage points (pp), while query pooling shows no clear advantage over matched mean pooling under balanced supervision.
\item \textbf{Evidence on representation use and compression.} Radar probes, token reductions, and an exact zero-weight control test transferred information, compression sensitivity, and a complete-to-subset prediction loss.
\end{enumerate}
Predictive dynamics and planning remain outside the evaluated scope.

\vspace{-2mm}
\section{Related Work}
\vspace{-1mm}
Multimodal learning joins representation, alignment, and fusion across sensor streams~\cite{baltrusaitis2019multimodal}, and models trained on complete observations can deteriorate when an input stream is unavailable~\cite{reza2024robust,ma2022multimodal}. Proposed responses include modality dropout~\cite{de2020input,neverova2015moddrop}, feature reconstruction and transformer fusion~\cite{woo2023good,lage2025physio}, shared and sensor-specific representations~\cite{wang2023shaspec}, logit distillation~\cite{wang2020distillation}, availability-conditioned paths~\cite{li2025simmlm}, and parameter-efficient adaptation~\cite{reza2024robust}. They do not address frozen JEPA representations of heterogeneous RGB, depth, and radar inputs evaluated across all seven subsets.

Joint-embedding predictive architectures predict representations rather than input samples, from I-JEPA for images~\cite{assran2023ijepa} to V-JEPA and V-JEPA2 for video~\cite{bardes2024vjepa,assran2025vjepa2}. Predictive alignment has also been studied across modalities and time series~\cite{lei2025m3jepa,girgis2024tsjepa,dutta2026charm}, while V-JEPA2-AC adds action-conditioned latent dynamics for robotic planning~\cite{assran2025vjepa2}. We instead use V-JEPA2 as a frozen feature extractor and test a complete-to-subset prediction objective against directly supervised fusion.

\vspace{-1mm}
\section{Proposed Methodology}
\vspace{-1mm}
All models use frozen V-JEPA2 features with the same fusion transformer and task heads, as summarised in Table~\ref{tab:comparators}, and the final query/all-view model uses the B3 inference architecture, as depicted in Figure~\ref{fig:architecture}(a).

\subsection{Sensor input construction}
\vspace{-1mm}
Let \(\modalities=\{\mathrm{RGB},\mathrm{D},\mathrm{R}\}\) denote the sensors and \(\emptyset\neq\visible\subseteq\modalities\) the available subset. RGB and depth inputs contain 64 frames at \(256\times256\) pixels. Depth is clipped at the 1st and 99th within-sample percentiles and scaled to the encoder range, and its frozen input convolution averages the pretrained colour kernels. Radar contributes 200 samples of phase displacement, filtered breathing, and filtered cardiac traces, which a complex Morlet wavelet transform maps to a stacked \(256\times256\) scalogram replicated over three channels and two identical frames. Every modality uses the frozen \texttt{vjepa2-vitl-fpc64-256} encoder without its predictor. \textcolor{black}{Encoded sequences are adaptively averaged to at most 128 tokens and cached.} Radar time lies along a scalogram axis, so duplicated frames do not establish temporal alignment with video tokens.

\subsection{Availability-aware fusion}
\vspace{-1mm}
\textcolor{black}{Query pooling uses a learned, modality-specific query to attend to the cached tokens and produce a 128-dimensional vector}, whereas mean pooling averages them before projection. The sensor slot is given by:
\vspace{-1mm}
\begin{equation}
\mathbf{s}_m=
\begin{cases}
\operatorname{pool}_m(\mathbf{T}_m)+\mathbf{e}_m+\mathbf{v}_1, &m\in\visible,\\
\mathbf{e}_m+\mathbf{v}_0,&m\notin\visible,
\end{cases}
\label{eq:slots}
\end{equation}
where \(\mathbf{T}_m\) holds the tokens, \(\mathbf{e}_m\) identifies the modality, and \(\mathbf{v}_0,\mathbf{v}_1\) encode absence and presence. A summary token and the three slots enter a pre-norm transformer with two layers, four heads, width 128, and feed-forward width 256. Its summary output \(\hrep_{\visible}\) feeds heads for three poses and four activities, whose decisions define 12 joint classes and whose mean cross-entropy is \(\mathcal{L}_{\mathrm{task}}\). Absent slots carry no reconstructed measurement.

\vspace{-3mm}
\subsection{Capacity and supervision controls}
\vspace{-1mm}
The factorial comparison crosses query and matched mean pooling with balanced and all-view supervision. Matched mean pooling adds a normalized residual \(128\rightarrow256\rightarrow128\) perceptron per sensor, giving \textcolor{black}{both downstream models 882,695 parameters} against 683,399 for Subset-mean (B2), without equalizing computation or inductive bias. Balanced supervision draws one subset per batch, \textcolor{black}{with selection counts across the seven subsets differing by at most one per epoch}. All-view supervision evaluates every subset on the same batch and makes one update with:
\vspace{-1mm}
\begin{equation}
\mathcal{L}_{\mathrm{all}}=\frac{1}{7}\sum_{\emptyset\ne S\subseteq\modalities}\frac{\mathcal{L}_{\mathrm{activity}}(S)+\mathcal{L}_{\mathrm{pose}}(S)}{2}.
\label{eq:allview}
\vspace{-1mm}
\end{equation}

The final model uses this objective alone, which gives seven supervised views per example and so increases exposure.

\vspace{-3mm}
\subsection{Complete-to-subset prediction}
\vspace{-1mm}
Consensus-JEPA adds a training-only complete-input branch. For \(\visible\subsetneq\modalities\), \(\zrep_{\visible}=q_\phi(g_\theta(\hrep_{\visible}))\) predicts \(\zrep_{\modalities}=\stopgrad[g_\theta(\hrep_{\modalities})]\) from the same online network, \textcolor{black}{where \(\stopgrad\) stops gradients through the complete-input target}. \textcolor{black}{The projector and predictor are two-layer perceptrons with hidden width 256 and output width 128. Training minimizes} \(\mathcal{L}_{\mathrm{task}}+\lambda\lossCJ\) with \(\lossCJ=1-\cos(\zrep_{\visible},\zrep_{\modalities})\). The complete pass runs in evaluation mode without a moving-average teacher. Zero-loss (A0) and Consensus (C1) execute the same graph with \(\lambda=0\) and \(0.003\), and both auxiliary modules are removed at inference.

\vspace{-1mm}

\section{Experimental Evaluation}
\label{sec:evaluation}
\vspace{-1mm}

\begin{table*}[ht!]
\def\arraystretch{1.0}
\setlength{\tabcolsep}{1.0em}
\centering
\caption{Earlier controlled reference models and their out-of-fold (OOF) recording-level macro-F1 (\%). All use frozen V-JEPA2 tokens, a fusion transformer, and supervised pose/activity heads, and universal models sample one subset per batch. Activity Avg-7 is mean $\pm$ sample SD, pose and joint are Avg-7 means, and Worst-7 averages the within-seed minimum.}
\label{tab:comparators}
\footnotesize
\input{tables_opus/tab_comparators_results}
\vspace{-4mm}
\end{table*}

\noindent\textbf{Dataset and acquisition protocol.} The complete aligned OMuSense-23 subset contains 39 subjects~\cite{omusense2026}, each recorded standing, sitting, and lying in four \textcolor{black}{nominally 30-second conditions} separated by 20-second gaps: normal breathing, reading aloud, guided breathing, and breath holding. The 117 sessions give 468 activity recordings. \textcolor{black}{Using TIMER-defined activity intervals, we extract 10-second windows at a nominal 1-second stride}, \textcolor{black}{appending a complete end-aligned window when the regular grid does not reach the interval end}. \textcolor{black}{This yields 10,530 aligned samples.} Lying-pose RGB and depth are rotated by \(180^\circ\) using the pose field, so deployment would need label-independent orientation metadata, and the camera position and background of this pose may affect pose and joint scores. Activity is therefore the primary endpoint, although acquisition effects cannot be excluded.


\vspace{1mm}
\noindent\textbf{Subject split, aggregation, and metrics.} Five outer subject-group folds each reserve five training subjects for validation, with subjects disjoint across training, validation, and testing. Recording-level activity macro-F1 selects checkpoints, on the training subset for Full-only and specialists and on the seven-subset mean for universal models. Window logits are averaged per recording before \(\hat y_r=\arg\max_c\bar\ell_{r,c}\), and the five test folds are pooled per seed, giving 468 predictions per subset. For task \(t\) with \(C_t\) classes,
\vspace{-1mm}
\begin{equation}
\begin{aligned}
F_{1,t}(S)&=\frac{1}{C_t}\sum_{c=1}^{C_t}\frac{2\,\mathrm{TP}_c}{2\,\mathrm{TP}_c+\mathrm{FP}_c+\mathrm{FN}_c},\\
\mathrm{Avg7}_{t}&=\frac{1}{7}\sum_{\emptyset\ne S\subseteq\modalities}F_{1,t}(S).
\end{aligned}
\label{eq:metrics}
\end{equation}
Activity Avg-7 is primary, with pose and joint Avg-7 and activity Worst-7, the within-seed minimum subset score. Paired 95\% intervals use 5,000 subject-cluster bootstrap draws shared between compared methods. They are conditional on the fitted models and folds, excluding retraining and new-cohort uncertainty, and exploratory without multiplicity correction.

\vspace{1mm}
\noindent\textbf{Comparators and optimization.} In Table~\ref{tab:comparators}, C1 must be compared with A0 to isolate the auxiliary weight, since B3 shares their inference architecture but not their training graph. Logit-KD targets complete-input logits of the same online network. These 14 configurations give 210 training tasks over five folds and three seeds, with AdamW at learning rate \(3\times10^{-4}\), weight decay \(10^{-4}\), batch size 128, gradient clipping at 1, at most 120 epochs, and patience 15. The factorial experiment keeps these settings, seeds 20260816--20260818, and validation Avg-7 selection, covering 60 training cells, 50 of them new.

\vspace{1mm}
\noindent\textbf{Representation probes and token reduction.} Linear probes fit separate logistic-regression heads (L-BFGS, \(C=1\), at most 2,000 iterations) to seven fixed representations, giving 70 heads over five folds without tuning or repeated seeds, with standardization fitted on training windows only. Radar uses a \(16\times16\) pooled scalogram, 135 spectral/statistical features (40 Hann-periodogram log-power bins over 0.1--4 Hz and five statistics per trace), mean V-JEPA2 features of 1,024 dimensions, and their 256-dimensional whitened PCA fitted on training data (seed 20260908). Mean RGB, mean depth, and full concatenation complete the set. Token reduction evaluates two 64-token variants of all 15 final checkpoints without retraining. The RGB/depth cache holds 32 temporal groups and four spatial stripes, so spatial aggregation gives \(32\times2\) groups and temporal aggregation \(16\times4\), while radar uses the same reduction in both.

\section{Results and Discussion}
\vspace{-1mm}
\noindent\textbf{Subset-aware training addresses the largest deficit.} Full-only reaches 80.51\% activity macro-F1 with all sensors but only 57.92\% Avg-7, whereas Subset-mean reaches 71.88\%, as shown in Table~\ref{tab:comparators}. This 13.96-point difference includes subset-aware checkpoint selection. It gains 31.69, 27.47, and 29.87 points for depth, radar, and depth+radar, but changes by -1.03 with complete input, so it mainly corrects the mismatch between training and available sensors. Frozen predictive features support the task, but the downstream decision must still be trained for the observed subsets.

\begin{figure*}[ht!]
\centering
\includegraphics[width=\linewidth]{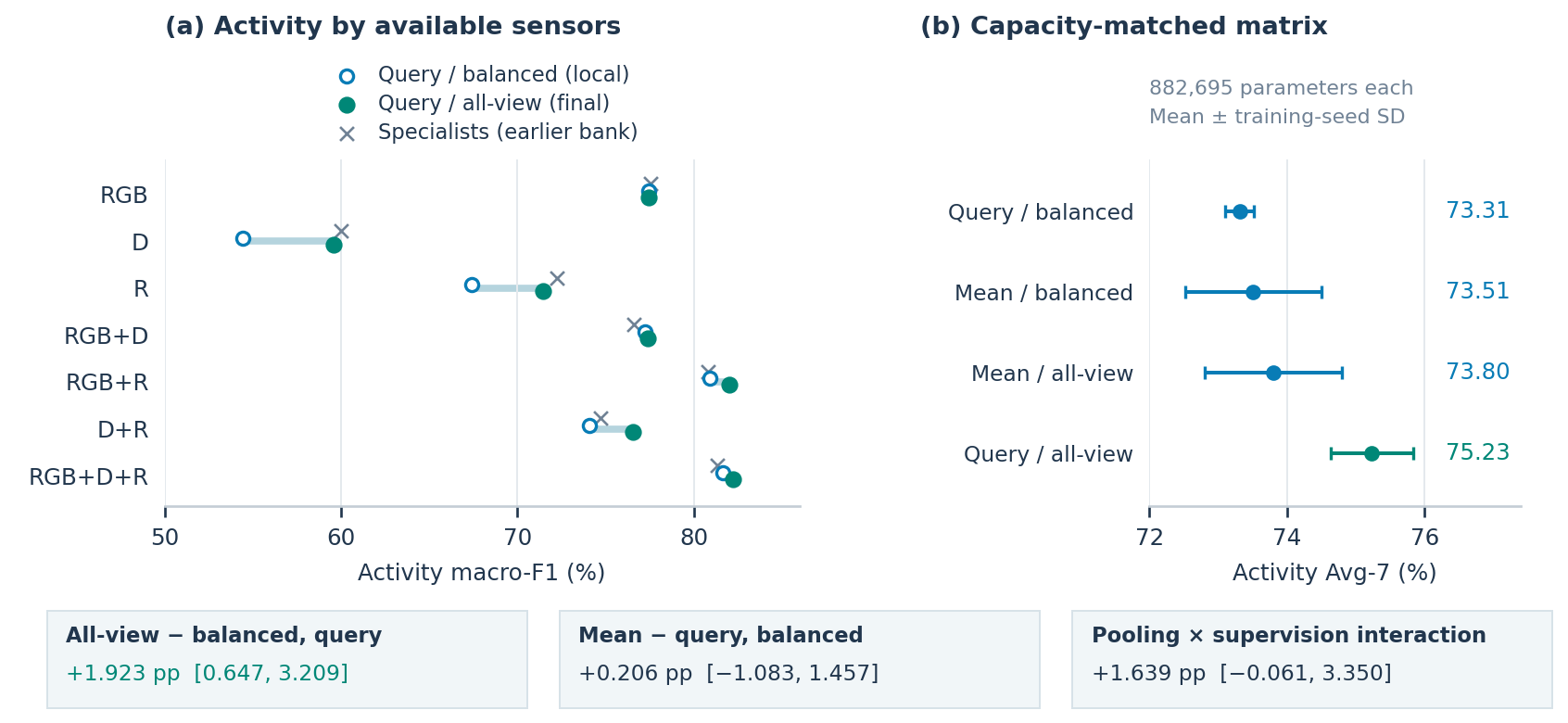}
\vspace{-6mm}
\caption{(a) Activity macro-F1 per sensor subset (three-seed means) for the local query/balanced control, the final query/all-view model, and the earlier specialists, with D depth and R radar. (b) Activity Avg-7 of the four capacity-matched models with training-seed SD, not subject intervals. Lower panels give paired contrasts in pp with 95\% subject-bootstrap intervals.}
\label{fig:results}
\end{figure*}

\vspace{1mm}
\noindent\textbf{Matched capacity qualifies the pooling claim.} The earlier 2.17 pp gain of Subset-query over Subset-mean also changes parameter count. With matched capacity, as shown in Table~\ref{tab:factorial} and depicted in Figure~\ref{fig:results}(b), mean minus query is +0.206 pp [\mbox{-1.083}, 1.457] under balanced supervision, while query exceeds mean by 1.433 pp [0.024, 2.879] under all-view supervision, close to the interval boundary. The interaction $(Q_{\mathrm{all}}-Q_{\mathrm{bal}})-(M_{\mathrm{all}}-M_{\mathrm{bal}})$ of 1.639 pp [\mbox{-0.061}, 3.350] does not establish a special synergy.

\begin{table}[ht!]
\def\arraystretch{1.0}
\setlength{\tabcolsep}{0.9em}
\centering
\caption{Completed capacity-matched comparison. All four models have 882,695 downstream parameters. Scores are recording-level macro-F1 (\%), with three-seed sample SD for activity Avg-7 and means for the remaining columns.}
\label{tab:factorial}
\footnotesize
\input{tables_opus/tab_factorial}
\end{table}

\vspace{1mm}
\noindent\textbf{All-view supervision improves the matched query model.} Query/all-view reaches $75.23\%\pm0.60$ activity Avg-7 against $73.31\pm0.21$ for its local query/balanced control, a paired gain of 1.923 pp [0.647, 3.209] that is positive for each seed. Pose and joint improve by 1.845 pp [1.069, 2.642] and 3.368 pp [1.919, 4.918]. The earlier B3 score of 74.05 records a different execution, not this same-environment reference. As depicted in Figure~\ref{fig:results}(a), the gain concentrates in incomplete inputs, with depth rising from 54.43 to 59.59\% and radar from 67.39 to 71.45\%. Depth remains the weakest subset, and the final model does not dominate every sensor condition.

\vspace{1mm}
\noindent\textbf{One network compared with seven specialists.} The final model is 0.490 pp [-0.607, 1.626] above the earlier specialist bank, which establishes neither superiority nor equivalence, and the bank was not rerun in the new environment. The specialists remain higher for depth and radar alone. Its 882,695 downstream parameters cover every availability state, against 4,381,745 for a bank pruned to its required branches, both excluding the shared frozen encoder of about 304 million parameters. This concerns storage and maintenance, not end-to-end speed or energy.

\vspace{1mm}
\noindent\textbf{Radar probes test representation transfer.} As shown in Table~\ref{tab:radar}, with the same classifier type, mean V-JEPA2 features exceed pooled CWT pixels by 6.319 pp [2.751, 9.996] in activity F1, and the PCA control retains 3.951 pp [0.287, 7.729] at the same input dimension. This supports transferred information beyond the pooled-image descriptor, although PCA fitting and pixel pooling prevent complete isolation of the encoder. The 2.143 pp [-1.217, 5.500] difference over spectral/statistical features leaves activity superiority uncertain. The radar probe also exceeds the radar-only shared model (76.31 against 71.45\%), so feature information is not equally used by every downstream configuration, although the gap does not identify interference.

\vspace{-2mm}
\begin{table}[ht!]
\def\arraystretch{1.1}
\setlength{\tabcolsep}{1.2em}
\centering
\caption{Fixed radar linear probes, with classifier input dimension and radar-only recording-level macro-F1 (\%), not seven-subset averages. Each row pools five OOF folds without repeated seeds.}
\label{tab:radar}
\footnotesize
\input{tables_opus/tab_radar}
\end{table}

\noindent\textbf{Post-training compression approximately preserves mean scores.} At 64 tokens the final checkpoints give activity Avg-7 of $75.26\%\pm0.34$ with spatial and $75.10\%\pm0.20$ with temporal aggregation. Spatial 64 minus the 128-token reference is +0.025 pp [-0.431, 0.484], and temporal minus spatial -0.152 pp [-0.589, 0.272], proving neither equivalence nor a preferred geometry, so 128 tokens remain the reference. The FP32 payload halves from 1.50 to 0.75 MiB per window, although the 128-token cache is still read. An earlier B3 reduction to 32 tokens lowered Avg-7 from 74.05 to 59.24\%. As reduction follows global encoder attention, neither experiment isolates local spatial or temporal information.

\vspace{1mm}
\noindent\textbf{The matched predictive loss does not improve Avg-7.} Consensus minus Zero-loss is -0.005 pp [-1.350, 1.278], a contrast that changes only the auxiliary weight. Direct alignment and Logit-KD do not exceed Subset-query, but change parameters or targets and do not show that such objectives generally fail. Over 15 Consensus checkpoints and six incomplete subsets, the median weighted auxiliary/task gradient-norm ratio is 0.065\%, and gradient cosine is negative in ten of 90 conditions. This indicates a weak auxiliary contribution at this weight, without establishing its cause or representation collapse, so the null result is specific to the tested target and protocol.

\vspace{1mm}
\noindent\textbf{What this contributes to JEPA-based world models.} The study separates transfer from a JEPA backbone, training an observation classifier for missing sensors, and learning predictive dynamics. This encoder could precede a separate transition model, but classification does not establish a physically sufficient or sensor-independent state. The transferable lesson is to test observation encoders across deployment subsets with controlled capacity, exposure, and auxiliary-loss baselines.

\vspace{1mm}
\noindent\textbf{Incremental evaluation and published sensing baselines.} Table~\ref{tab:stages} places our stages beside published models that process engineered camera/radar or derived cardiorespiratory signals~\cite{omusense2026,lage2025physio,lage2025hybrid}. The initial distillation used an RGB--radar teacher and rolling leave-one-subject-out evaluation with three validation subjects and window metrics, whereas the controlled stage adds subject-group folds, recording aggregation, and corrected orientation and radar windowing. \textcolor{black}{Because protocols differ, 75.23\% Avg-7 cannot be interpreted as a controlled improvement over 70.43\%. Likewise, the full-input 82.23\% macro-F1 and published 87.2\% hybrid accuracy cannot establish a performance ranking because both the metrics and evaluation protocols differ.}

\begin{table}[ht!]
\def\arraystretch{1.0}
\setlength{\tabcolsep}{1.0em}
\centering
\caption{Published OMuSense-23 baselines and stages of this work, as activity accuracy/macro-F1 (\%). Published protocols are subject fivefold, except the 30/10/10 split of physiological SSL. Stages are not matched reruns, earlier F1 aggregation differs from pooled OOF scoring, and a dash marks an untabulated result.}
\label{tab:stages}
\footnotesize
\input{tables_opus/tab_stages}
\end{table}


\vspace{1mm}
\noindent\textbf{Scope and remaining evidence.} We test declared absence, not failure detection or corrupted inputs, and all seven subsets appear in training. Unseen sensors, incomplete records, free-living activity, new sites, and other backbones are not evaluated. Training compute and specialist exposure are unmatched, pose may use acquisition context, and repeated development on this cohort limits the intervals. An exposure control with one update from seven independent batches differs from all-view by +0.140 and -0.488 pp in two validation cells, so a unique benefit from reusing examples is not established. \textcolor{black}{Saved predictions and paired intervals are available. Full reproduction also requires model checkpoints, cached features, and subject splits.} Other small fusion and auxiliary-loss screens remain exploratory.

\section{Conclusion}
We present a shared multimodal fusion model that recognizes breathing activity and pose from frozen V-JEPA2 representations of RGB, depth, and radar under all seven non-empty sensor subsets, without reconstructing missing measurements. Supervising all subsets in each batch yields an activity Avg-7 of $75.23\%\pm0.60$ and improves the matched control by 1.923 pp, primarily for incomplete inputs. Under matched capacity, query pooling shows no clear advantage over matched mean pooling under balanced supervision, and the Consensus-JEPA loss yields no measurable gain. The evidence therefore supports direct supervision over the known availability states with explicit checks of representation, capacity, uncertainty, and compression, whereas predictive state modelling for JEPA-based world models remains a separate question.


\section*{\centering\normalsize ACKNOWLEDGMENT}
The research was supported by the Research Council of Finland Smart Video Sensorization for Secure Healthcare Monitoring (SViSenS) (Grant Number: $370277$), the Interreg Aurora ResilientEdge project (Grant Number: 20373282) and Profi7 Hybrid intelligence program (352788). The authors wish to acknowledge the use of AI-assisted tools for language editing, with a focus on grammar checking and improving readability, and coding support. In addition, the Elicit platform was used during the literature search as an AI-assisted discovery aid for candidate records. All inclusion decisions, exclusions, data extraction, synthesis, interpretation, and final writing decisions were performed by the authors.

\section*{\centering\normalsize COMPLIANCE WITH ETHICAL STANDARDS}
This work analyses existing recordings, including RGB video, from the OMuSense-23 acquisition~\cite{omusense2026}. That study was conducted in accordance with the Declaration of Helsinki and institutional regulations, and all participants provided written informed consent. No new data were collected and no participant intervention was performed. The face shown in Figure~\ref{fig:architecture} is pixelated to prevent identification.

\bibliographystyle{IEEEbib}
\bibliography{refs_v2}

\end{document}

%% file: tables_opus/tab_comparators_results.tex
\begin{tabular}{@{}lcllllrrr@{}}
\toprule
Model (ID) & Nets & Training subsets & Pooling & Auxiliary target (weight) & Act. Avg-7 & Pose & Joint & Worst-7 \\
\midrule
Full-only (B0) & 1 & Full input only & Mean & None & $57.92\pm2.67$ & 78.99 & 48.82 & 22.35 \\
Subset-mean (B2) & 1 & Balanced seven & Mean & None & $71.88\pm0.74$ & 89.80 & 64.15 & 54.04 \\
Subset-query (B3) & 1 & Balanced seven & Query & None & $74.05\pm0.36$ & 90.09 & 66.64 & 55.22 \\
Specialists (B1) & 7 & One fixed subset & Query & None & $74.74\pm0.55$ & 90.29 & 67.63 & 60.00 \\
\midrule
Zero-loss (A0) & 1 & Balanced seven & Query & Full latent branch (0) & $73.24\pm1.56$ & 90.07 & 65.92 & 54.26 \\
Consensus (C1) & 1 & Balanced seven & Query & Full latent, predictor (0.003) & $73.23\pm1.16$ & 90.06 & 65.91 & 54.73 \\
Direct (A1) & 1 & Balanced seven & Query & Full latent, no predictor (0.003) & $73.69\pm0.82$ & 89.82 & 66.08 & 54.93 \\
Logit-KD (KD) & 1 & Balanced seven & Query & Full logits (0.5, temperature 2) & $73.81\pm1.32$ & 89.91 & 66.55 & 54.49 \\
\bottomrule
\end{tabular}

%% file: tables_opus/tab_factorial.tex
\begin{tabular}{@{}lrrrr@{}}
\toprule
Pooling / supervision & Act. Avg-7 & Pose & Joint & Worst-7 \\
\midrule
Query / balanced & $73.31\pm0.21$ & 89.41 & 65.49 & 54.43 \\
Mean / balanced & $73.51\pm0.99$ & 89.61 & 65.58 & 56.09 \\
Mean / all-view & $73.80\pm1.00$ & 90.47 & 66.72 & 56.70 \\
\textbf{Query / all-view} & $\mathbf{75.23\pm0.60}$ & 91.25 & 68.85 & 59.59 \\
\bottomrule
\end{tabular}

%% file: tables_opus/tab_radar.tex
\begin{tabular}{@{}lrrrr@{}}
\toprule
Radar representation & Dim. & Activity & Pose & Joint \\
\midrule
Pooled CWT pixels & 256 & 69.99 & 71.30 & 45.06 \\
Spectral/statistical & 135 & 74.17 & 76.00 & 52.91 \\
V-JEPA2 mean + PCA & 256 & 73.95 & 82.22 & 60.49 \\
V-JEPA2 mean & 1,024 & 76.31 & 84.58 & 64.17 \\
\bottomrule
\end{tabular}

%% file: tables_opus/tab_stages.tex
\begin{tabular}{@{}lcc@{}}
\toprule
Study/stage (representation) & Full input & Seven subsets \\
\midrule
\multicolumn{3}{@{}l}{\textit{Published, $N=50$, window-level protocols}} \\
OMuSense-23~\cite{omusense2026} (477 handcrafted) & 81.0/81.0 & -- \\
Physiological SSL~\cite{lage2025physio} (four 1-D) & 74.0/-- & -- \\
Hybrid fusion~\cite{lage2025hybrid} (handcrafted+deep) & 87.2/-- & -- \\
\midrule
\multicolumn{3}{@{}l}{\textit{Initial stage, $N=39$, rolling LOSO/window}} \\
V-JEPA2 (frozen RGB, depth, radar) & 77.99/76.27 & -- \\
Distillation (experts + teacher) & 81.33/79.80 & 72.39/70.43 \\
\midrule
\multicolumn{3}{@{}l}{\textit{Controlled stage, $N=39$, fivefold OOF/recording}} \\
Full-only (B0), corrected cache & --/80.51 & --/57.92 \\
Subset-query (B3), same cache & --/81.18 & --/74.05 \\
Query/all-view, final & --/82.23 & --/75.23 \\
\bottomrule
\end{tabular}

%% file: refs.bib
@inproceedings{wang2023shaspec,
  title = {Multi-modal Learning with Missing Modality via Shared-Specific Feature Modelling},
  author = {Wang, Hu and Chen, Yuanhong and Ma, Congbo and Avery, Jodie and Hull, Louise and Carneiro, Gustavo},
  booktitle = {Proceedings of the IEEE/CVF Conference on Computer Vision and Pattern Recognition},
  year = {2023},
  eprint = {2307.14126},
  archivePrefix = {arXiv}
}

@misc{neverova2015moddrop,
  title = {ModDrop: Adaptive Multi-Modal Gesture Recognition},
  author = {Neverova, Natalia and Wolf, Christian and Taylor, Graham W. and Nebout, Florian},
  year = {2015},
  eprint = {1501.00102},
  archivePrefix = {arXiv}
}

@inproceedings{li2025simmlm,
  title={Simmlm: A simple framework for multi-modal learning with missing modality},
  author={Li, Sijie and Chen, Chen and Han, Jungong},
  booktitle={Proceedings of the IEEE/CVF International Conference on Computer Vision},
  pages={24068--24077},
  year={2025}
}


%% file: refs_v2.bib
@article{omusense2026,
  author = {Lage Ca{\~n}ellas, Manuel and Nguyen, Le and Mukherjee, Anirban and
            {\'A}lvarez Casado, Constantino and Wu, Xiaoting and Nguyen, Nhi and
            Susarla, Praneeth and Sharifipour, Sasan and Jayagopi, Dinesh B. and
            Bordallo L{\'o}pez, Miguel},
  title = {{OMuSense-23}: A Multimodal Dataset for Contactless Breathing Pattern
           Recognition and Biometric Analysis},
  journal = {IEEE Internet of Things Journal},
  pages = {1--14},
  year = {2026},
  doi = {10.1109/JIOT.2026.3720896}
}

@article{baltrusaitis2019multimodal,
  author = {Baltru{\v{s}}aitis, Tadas and Ahuja, Chaitanya and Morency,
            Louis-Philippe},
  title = {Multimodal Machine Learning: A Survey and Taxonomy},
  journal = {IEEE Transactions on Pattern Analysis and Machine Intelligence},
  volume = {41},
  number = {2},
  pages = {423--443},
  year = {2019},
  doi = {10.1109/TPAMI.2018.2798607}
}

@inproceedings{woo2023good,
  author = {Woo, Sangmin and Lee, Sumin and Park, Yeonju and Nugroho,
            Muhammad Adi and Kim, Changick},
  title = {Towards Good Practices for Missing Modality Robust Action Recognition},
  booktitle = {Proceedings of the AAAI Conference on Artificial Intelligence},
  volume = {37},
  pages = {2776--2784},
  year = {2023},
  doi = {10.1609/aaai.v37i3.25378}
}

@inproceedings{wang2020distillation,
  author = {Wang, Qi and Zhan, Liang and Thompson, Paul and Zhou, Jiayu},
  title = {Multimodal Learning with Incomplete Modalities by Knowledge
           Distillation},
  booktitle = {Proceedings of the 26th ACM SIGKDD International Conference on
               Knowledge Discovery and Data Mining},
  year = {2020},
  doi = {10.1145/3394486.3403234}
}

@inproceedings{assran2023ijepa,
  author = {Assran, Mahmoud and Duval, Quentin and Misra, Ishan and Bojanowski,
            Piotr and Vincent, Pascal and Rabbat, Michael and LeCun, Yann and
            Ballas, Nicolas},
  title = {Self-Supervised Learning from Images with a Joint-Embedding
           Predictive Architecture},
  booktitle = {Proceedings of the IEEE/CVF Conference on Computer Vision and
               Pattern Recognition},
  pages = {15619--15629},
  year = {2023}
}

@article{bardes2024vjepa,
  author = {Bardes, Adrien and Garrido, Quentin and Ponce, Jean and Chen, Xinlei
            and Rabbat, Michael and LeCun, Yann and Assran, Mahmoud and Ballas,
            Nicolas},
  title = {Revisiting Feature Prediction for Learning Visual Representations
           from Video},
  journal = {arXiv preprint arXiv:2404.08471},
  year = {2024},
  doi = {10.48550/arXiv.2404.08471}
}

@article{assran2025vjepa2,
  author = {Assran, Mido and Bardes, Adrien and Fan, David and Garrido, Quentin
            and Howes, Russell and Komeili, Mojtaba and Muckley, Matthew and
            Rizvi, Ammar and Roberts, Claire and Sinha, Koustuv and others},
  title = {{V-JEPA 2}: Self-Supervised Video Models Enable Understanding,
           Prediction and Planning},
  journal = {arXiv preprint arXiv:2506.09985},
  year = {2025},
  doi = {10.48550/arXiv.2506.09985}
}

@inproceedings{lei2025m3jepa,
  author = {Lei, Hongyang and Cheng, Xiaolong and Qin, Qi and Wang, Dan and
            Huang, Huazhen and Gu, Qingqing and Wu, Yetao and Ji, Luo},
  title = {{M3-JEPA}: Multimodal Alignment via Multi-Gate {MoE} Based on the
           Joint-Embedding Predictive Architecture},
  booktitle = {Proceedings of the 42nd International Conference on Machine
               Learning},
  series = {Proceedings of Machine Learning Research},
  volume = {267},
  pages = {33902--33917},
  year = {2025}
}

@article{girgis2024tsjepa,
  author = {Girgis, Abanoub M. and Valcarce, Alvaro and Bennis, Mehdi},
  title = {Time-Series {JEPA} for Predictive Remote Control under
           Capacity-Limited Networks},
  journal = {arXiv preprint arXiv:2406.04853},
  year = {2024},
  doi = {10.48550/arXiv.2406.04853}
}

@article{dutta2026charm,
  author = {Dutta, Utsav and Pastrana, Gerardo and Khoshfetrat Pakazad, Sina and
            Ohlsson, Henrik},
  title = {Giving Sensors a Voice: Multimodal {JEPA} for Semantic Time-Series
           Embeddings},
  journal = {arXiv preprint arXiv:2605.31580},
  year = {2026},
  doi = {10.48550/arXiv.2605.31580}
}

@article{lage2025physio,
  author = {Lage Ca{\~n}ellas, Manuel and {\'A}lvarez Casado, Constantino and
            Nguyen, Le and Bordallo L{\'o}pez, Miguel},
  title = {A Self-Supervised Multimodal Framework for {1D} Physiological Data
           Fusion in Remote Health Monitoring},
  journal = {Information Fusion},
  volume = {124},
  pages = {103397},
  year = {2025},
  doi = {10.1016/j.inffus.2025.103397}
}

@inproceedings{lage2025hybrid,
  author = {Lage Ca{\~n}ellas, Manuel and {\'A}lvarez Casado, Constantino and
            Malin, Miika and Prencipe, Nicoletta and Sharifipour, Sasan and
            Bordallo L{\'o}pez, Miguel},
  title = {Fusion of Handcrafted and Deep-Learned Cardiorespiratory Features
           for Breathing Pattern Classification},
  booktitle = {Proceedings of the 33rd European Signal Processing Conference
               (EUSIPCO)},
  pages = {1627--1631},
  address = {Palermo, Italy},
  publisher = {IEEE Signal Processing Society},
  year = {2025},
  isbn = {978-9-46-459362-4},
  url = {https://eusipco2025.org/wp-content/uploads/pdfs/0001627.pdf}
}

@article{reza2024robust,
  title={Robust Multimodal Learning With Missing Modalities via Parameter-Efficient Adaptation},
  author={Reza, Md Kaykobad and Prater-Bennette, Ashley and Asif, M. Salman},
  journal={IEEE Transactions on Pattern Analysis and Machine Intelligence},
  volume={47},
  number={2},
  pages={742--754},
  year={2025},
  doi={10.1109/TPAMI.2024.3476487}
}

@inproceedings{de2020input,
  title={Input dropout for spatially aligned modalities},
  author={de Blois, S{\'e}bastien and Garon, Mathieu and Gagn{\'e}, Christian and Lalonde, Jean-Fran{\c{c}}ois},
  booktitle={2020 IEEE International Conference on Image Processing (ICIP)},
  pages={733--737},
  year={2020},
  organization={IEEE}
}

@inproceedings{ma2022multimodal,
  title={Are multimodal transformers robust to missing modality?},
  author={Ma, Mengmeng and Ren, Jian and Zhao, Long and Testuggine, Davide and Peng, Xi},
  booktitle={2022 IEEE/CVF conference on computer vision and pattern recognition (CVPR)},
  pages={18156--18165},
  year={2022},
  organization={IEEE}
}
